\pdfoutput=1

\documentclass[11pt]{article}

\usepackage{emnlp2021}

\usepackage{times}
\usepackage{latexsym}

\usepackage[T1]{fontenc}

\usepackage[utf8]{inputenc}

\usepackage{microtype}

\usepackage{subcaption}
\usepackage[disable]{todonotes}

\usepackage{siunitx}
\usepackage{amsfonts}
\usepackage{amsmath}
\usepackage{amssymb}

\title{Humans and language models diverge when predicting repeating text}

\author{Aditya R. Vaidya \\
  UT Austin \\
  \texttt{avaidya@utexas.edu} \\\And
  Javier Turek \\
  Intel Labs  \\
  \texttt{javier.turek@intel.com} \\\And
  Alexander G. Huth \\
  UT Austin \\
  \texttt{huth@cs.utexas.edu} \\}

\begin{document}
\maketitle

\begin{abstract}
Language models that are trained on the next-word prediction task have been shown to accurately model human behavior in word prediction and reading speed. In contrast with these findings, we present a scenario in which the performance of humans and LMs diverges. We collected a dataset of human next-word predictions for five stimuli that are formed by repeating spans of text. Human and GPT-2 LM predictions are strongly aligned in the first presentation of a text span, but their performance quickly diverges when memory (or in-context learning) begins to play a role. We traced the cause of this divergence to specific attention heads in a middle layer. Adding a power-law recency bias to these attention heads yielded a model that performs much more similarly to humans. We hope that this scenario will spur future work in bringing LMs closer to human behavior.\footnote{Data and code are publicly available at: \url{https://github.com/HuthLab/lm-repeating-text}}
\end{abstract}

\section{Introduction}

Transformer-based language models (LMs) are neural networks that are trained to predict upcoming words from their preceding context. These models flexibly retrieve and combine information across a context that might span thousands of words, enabling them to learn from in-context examples~\cite{daiWhyCanGPT2022,xieExplanationIncontextLearning2022,olssonIncontextLearningInduction2022}, tell coherent stories~\cite{leeCoAuthorDesigningHumanAI2022}, and perform many other advanced language tasks~\cite{tiedemannOPUSMTBuildingOpen2020,brownLanguageModelsAre2020}. 

These abilities far surpass any previous computational models or linguistic theories~\cite{yangOneModelLearning2022}, leading many to use LMs as models of human cognition. For example, LM surprisal---a measure of how well it can predict the next word---has been found to be highly correlated with both how long humans spend reading each word~\cite{goodkindPredictivePowerWord2018,haoProbabilisticPredictionsPeople2020,wilcoxPredictivePowerNeural2020} and the accuracy of human next-word predictions~\cite{goldsteinThinkingAheadSpontaneous2021,jacobsHumanUnlikenessNeural2020}. These results suggest that LMs and humans might be using similar mechanisms to structure and recall information from memory. However, these seeming parallels have not gone unchallenged. \citet{ohWhyDoesSurprisal2023}, for example, showed that LM surprisal and human reading time become decorrelated as models grow in size and power, suggesting a more superficial relationship than previously thought. 

In this work we test whether apparent similarities between LM and human next-word prediction accuracy reflect true similarities in memory mechanisms.
To accomplish this we introduce a new task that combines memory with next-word prediction using repeating natural text stimuli. Comparing human behavioral performance with an LM, we found that LM surprisal \textit{decorrelates} from human predictions in this scenario.
While human performance improves modestly with each repetition, the transformer-based LM GPT-2 \cite{radfordLanguageModelsAre2019} reaches near-perfect performance after just one presentation.
To better understand this behavior, we examined the patterns of memory access (via attention) in the model, revealing how the model solves this task. We then showed that the model can be made to perform more like the humans by adjusting these patterns to mimic human memory~\cite{donkinPowerLawModelPsychological2012}.

This work demonstrates an important way in which human and LM memory mechanisms diverge, casting doubt on the use of existing LMs as a model of human cognition. However, the framework we developed for making the model more human-like also provides a potential way forward. Directly optimizing LMs for human-like behavior---including but not limited to memory tasks like that used here---could lead to much better computational models of human cognition and memory. It is also possible that investigating the relationship between human and model memory could provide guidance for developing better, more efficient neural network models.

\begin{figure*}[h!]
    \centering
    \includegraphics[width=0.75\textwidth]{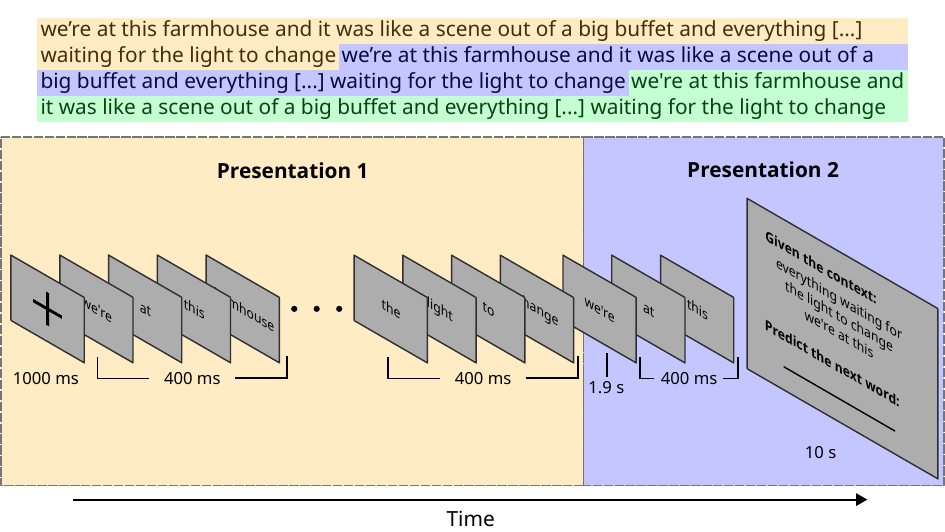}
    \caption{Paradigm for collecting human next-word predictions. A span of text is presented three times without break. Each presentation of the stimulus is denoted with a different color. Subjects are shown words one-at-a-time with RSVP. When prompted to predict the next word, subjects are shown the previous 10 words and are given 10 seconds to type their prediction. After submitting a response, presentation of the stimulus resumes. If incorrect, they are first shown the correct word and must acknowledge before continuing.}
    \label{fig:paradigm}
\end{figure*}

\section{Related works}
Human performance on recall tasks, like the experiment we propose here, is primarily limited by short-term memory~\cite{baddeleyWorkingMemory1992}.
In these tasks, humans show both recency biases (i.e. better recall for the most recent items) and primacy biases (better for the first items)~\cite{tzengPositiveRecencyEffect1973,jefferiesAutomaticControlledProcessing2004}.
Recall tasks often show repetition effects; presenting a stimulus multiple times successively decreases the recall error rate~\cite{kintschEffectsRepetitionShortterm1965,baddeleyReactionTimeShortterm1973,amlundRepetitiveReadingRecall1986}.
Some have suggested a link between language deficits and the number of presentations needed to reach perfect verbatim sentence recall~\cite{milesVerbatimGistRecall2006}.
Many studies have also shown that human memory decay follows a power law~\cite{donkinPowerLawModelPsychological2012}, where, for example, the number of items accurately recalled from a list will decrease over time $t$ proportional to $t^{-d}$ for some constant decay rate $d$.

Transformers neural networks, in contrast with humans, can attend to exact token identities hundreds or thousands of tokens in the past at no additional cost, subject only to the context length.
One limitation of the standard attention implementation is that memory and runtime scale quadratically with the number of tokens, making longer inputs prohibitively expensive.
Recently, significant work has gone into extending the maximum context length for transformers while avoiding these computational issues.
Transformer-XL caches hidden states to allow attention to tokens beyond the immediate input~\cite{daiTransformerXLAttentiveLanguage2019}.
FlashAttention is an optimized attention algorithm that exploits the hardware architecture to train models with context lengths up to 64K tokens~\cite{daoFlashAttentionFastMemoryEfficient2022}.
The ALiBi method~\cite{pressTrainShortTest2022} replaces sinusoidal positional embeddings with a recency bias on the attention scores, such that closer query-key pairs are weighted higher than more distant pairs. Using ALiBi necessitates retraining a model with the new attention mechanism, though once trained it can generalize to longer lengths.

\begin{figure*}[ht]
    \centering
    \includegraphics[]{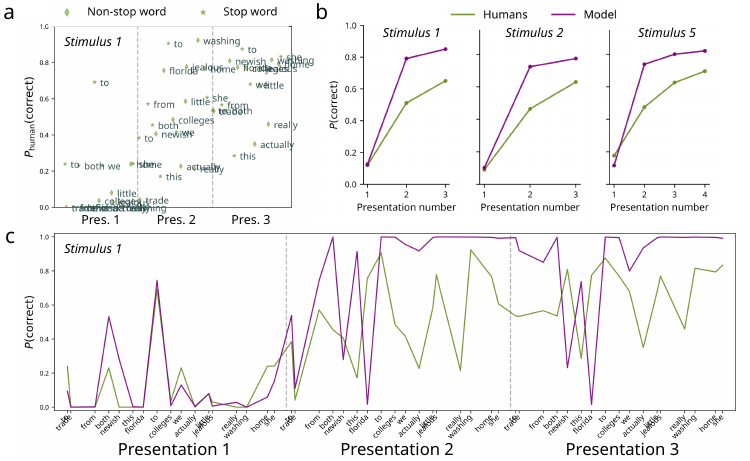}
    \caption{Behavioral and model results. (a) Human next-word prediction accuracy for one stimulus. Prompted words are split into stop words and non-stop words using the stop word list from NLTK~\cite{BirdKleinLoper09}. Dotted vertical lines indicate the boundaries between presentations. (b) Human and model performance, averaged within each presentation, for three different stimuli. Stimuli 1 and 2 were presented three times, while Stimulus 5 was presented four times. Both model and human accuracy improve over presentations, but model performance improves much faster and reaches a higher level. (c) Timecourse for human (green) and model (purple) performance for the stimulus from (a).}
    \label{fig:basic_results}
\end{figure*}

\section{Human behavioral study}

We first designed an experiment to evaluate human memory in a next-word prediction task with repeated word sequences.
We then compared the humans against an LM on the same stimuli to evaluate the LM's memory.

\subsection{Setup for humans}

We collected human next-word predictions on repeating stimuli from a corpus of spoken story transcripts~\cite{lebelNaturalLanguageFMRI2023}.
To construct the stimuli, we chose five phrase-aligned spans of between 40 and 100 words (without punctuation) from the corpus and repeated each span between one and three times, for a total of between 2 and 4 presentations of the span.
One span was repeated once; three spans were repeated twice; and one span was repeated three times.
The stimuli can be seen in Section~\ref{sec:appendix_stimuli} in the Appendix.
Subjects were presented words one-at-a-time via rapid serial visual presentation (RSVP; \citealp{potterRapidSerialVisual1984}) at a fixed duration of \SI{400}{\ms} per word, with \SI{1.5}{\s} pauses at the end of each presentation.
At predetermined moments, subjects were prompted to predict the next word given the previous 10 words. Prompts appeared roughly every 13 words, giving the subjects time to process the story naturally between interruptions.
Figure~\ref{fig:paradigm} shows the presentation of the stimuli and an example prompt screen.

To ensure that we could measure memory effects robustly, 50\% of a given subject's prompts were at the same position in all presentations of a stimulus, while the other 50\% were only prompted on a single presentation.
Within each presentation, prompts were selected by taking a weighted random sample of the words to provide a balanced selection of low- and high-frequency words.
Weights were calculated as the average of two values: the complement of the unigram probability and the reciprocal of the unigram probability.
Both weights were normalized to sum over words to 1 before being averaged.
Subjects were told at the beginning of the experiment that the word sequences will repeat, but were not told where.
Human performance $P_{\text{human}}(\text{correct})$ was calculated as the proportion of participants whose responses exactly match the ground-truth next word, ignoring case and leading or trailing whitespace.

In total, 100 online participants were recruited through Prolific (\url{www.prolific.co}). Subjects were required to be fluent in English and were given performance-based bonus compensation.
The online experiment was constructed using the Gorilla Experiment Builder (\url{www.gorilla.sc}).
The experimental protocol was approved by the Institutional Review Board at The University of Texas at Austin.
Written consent was obtained from all subjects.

\begin{figure*}
    \centering
    \includegraphics[width=\textwidth]{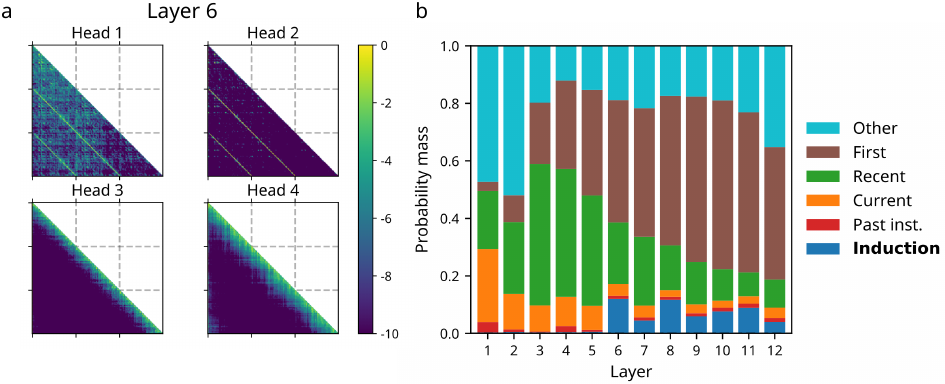} 
    \caption{Attention patterns. (a) Attention matrices for four heads in layer 6 for Stimulus 1 (65-word span presented 3 times). Plotted is the log-attention. Dotted gray lines indicate boundaries between presentations. Strong diagonals demonstrating induction from previous presentations are present in heads 1 and 2, but not 3 and 4. (b) Summarized attention patterns across layers. Probability mass of each category is averaged across all tokens, all heads for the given layer, and all stimuli. Induction-like attention emerges sharply at layer 6 and is present in each subsequent layer.}
    \label{fig:attention}
\end{figure*}

\subsection{Setup for language models}

We used a pre-trained GPT-2 Small~\cite{radfordLanguageModelsAre2019} model, which we fine-tuned to change its tokenization from BPE~\cite{sennrichNeuralMachineTranslation2016} to word-level (i.e., whitespace-delimited) so that its tokenization scheme would match the experimental protocol for the human participants. 
We used non-repeating story transcripts as training data for fine-tuning and excluded the stories used to construct the behavioral stimuli.
To get model prediction probabilities for comparison with the human data, we fed the entire repeating stimulus into the model and calculated the top-1 accuracy for each token.

\section{Behavioral study results}

Figure~\ref{fig:basic_results}a shows human performance on one text span; as they are shown more words, human accuracy generally increases.
Many stop words are predicted well even during the first presentation, while non-stop words improve more linearly with the number of presentations.
Humans consistently improve as they are shown more presentations of the same text span, as seen in Figure~\ref{fig:basic_results}b. While the model accuracy is similar to humans on the first presentation, it quickly jumps to a much higher level thereafter.

A more detailed view appears in Figure~\ref{fig:basic_results}c, where we show accuracy for both model and human on each probe word. GPT-2 accuracy is strongly correlated with human accuracy for the initial presentation of this span ($r=0.87$), replicating earlier findings~\cite{goldsteinThinkingAheadSpontaneous2021}. However, model and human accuracies markedly diverge thereafter, with the correlation dropping to $r=0.24$ in the second presentation and $r=0.05$ in the third.

These results provide a potent counterexample to previous claims of alignment: Humans and LMs only seem to behave similarly in the initial presentation of a stimulus, but produce uncorrelated behavior once short-term memory comes into play. %
This suggests that the model and humans are exploiting very different memory mechanisms to solve this task. The humans must rely on lossy short-term memory, while the model can leverage in-context learning to provide super-human, near-perfect recall. 
While earlier reports suggested that such detailed recall might mimic human working memory~\cite{armeniCharacterizingVerbatimShortTerm2022}, these results suggest that the models go well beyond human capabilities.

\section{Patterns in model attention}
Our behavioral results show that human and LM next-word prediction diverge sharply when short-term memory is involved, suggesting that the two systems use substantially different memory mechanisms. To gain insight into the cause of these differences, we next sought to understand how exactly the model was able to achieve such high performance on this task.

``Memory'' in transformer models is implemented by using dot-product attention over previous words. Each of the 12 layers in this model contains 12 attention heads, each of which looks for specific features in the content or location of previous words. The action of each attention head can be summarized in an \textit{attention matrix}, $A$, which shows how much attention token $i$ is paying to token $j$ for all $j<i$. Attention weights are normalized so that each row $A_i$ of the attention matrix sums to 1. The values in the attention matrix can thus show us how and where the model is ``recalling'' past information.

Previous work on simplified transformer models has identified the emergence of specific attention heads that recognize patterns in the input and produce outputs that complete those patterns~\cite{elhageMathematicalFrameworkTransformer2021,olssonIncontextLearningInduction2022}.
These \textit{induction heads} specifically attend to the token after the previous presentation of the current (input) token, essentially allowing the model to read out the completion from a previous instance of the same pattern.
For inputs that are constructed from repeating sequences---like those used in our behavioral experiment---induction heads should thus produce a highly stereotypical attention matrix:
If a stimulus consists of repeating spans of length $k$, the head attends to the token $k-1$ tokens in the past.

We examined the attention matrices of GPT-2 Small for our stimuli and found multiple heads across many layers that exhibit induction behavior.
Figure~\ref{fig:attention}a depicts example attention matrices for four heads in layer 6. While attention values are non-negative and sum to 1 in each row, we use log-scaled values here to highlight subtle effects.
For this test the stimulus consisted of three presentations of a $65$-word span, so an induction head should attend to the word appearing $64$ positions ago, which is exactly the word that the model should output at each point. This should manifest as strong diagonals in the attention matrix.
This is exactly the pattern that we see for attention heads 1 and 2.
Further, when processing tokens in the third presentation, these heads attend to previous instances in both of the first two presentations ($64$ and $129$ tokens in the past).
To illustrate that this pattern is not found everywhere in the model, we also show two other attention heads (3 and 4) from the same layer, which exhibit no induction-like behavior, but instead attend to recent words.

To more efficiently find induction-like behavior in the model, we can summarize how well the attention matrix for each head matches a few different patterns.
For each layer, we quantified the average probability mass attributable to the heads attending to:
\begin{itemize}
    \item the first token in the input, often thought to represent a sort of ``default'' attention state~\cite{olssonIncontextLearningInduction2022},
    \item the 5 most recent tokens (likely capturing local syntactic effects),
    \item the current token,
    \item past instances of the current token,
    \item the token \textit{after} each past instance of the current token (induction), and
    \item all other tokens.
\end{itemize}
Figure~\ref{fig:attention}b shows the probability mass given to each attention pattern in each layer, averaged across all 12 heads.
We see that the induction attention pattern arises sharply and specifically in layer 6 and continues through the output layer (layer 12).
These results suggest that these layers---and especially layer 6---have a causal role in copying words from previous repetitions of the text span, and thus may be the source of the divergence in human-LM accuracy.
In the next section, we test this hypothesis by selectively disrupting each layer in an attempt to make the model more human-like.

\begin{figure*}[ht]
    \centering
    \includegraphics[width=\textwidth]{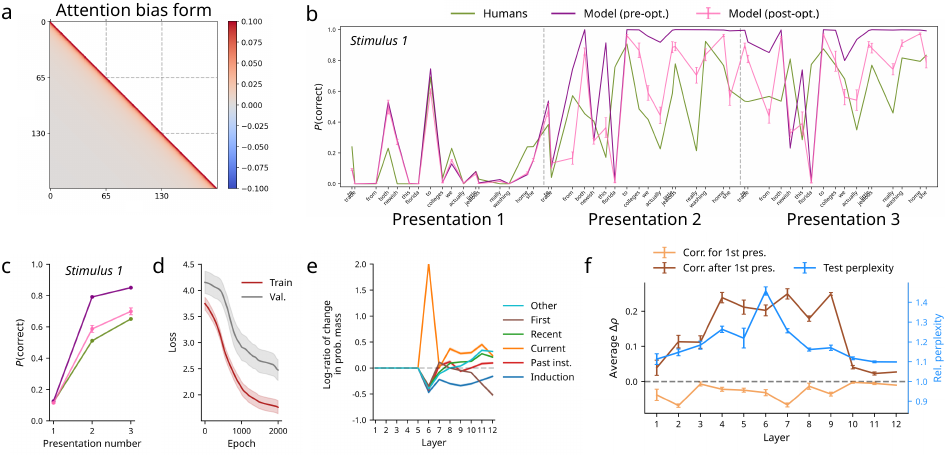} 
    \caption{Attention bias optimization. (a) An example bias matrix that would give the attention head a recency bias ($\alpha_{h} = 0.373, \beta_{h} = 0.0049$). (b) Example timecourse that shows human performance (green), original model performance (purple), and post-optimization held-out model performance (pink). Error bars indicate SEM across initializations. (c) Human and model performance, averaged within presentations, for the same stimulus. (d) Average training and validation curves. The validation curve is the MSE on a randomly selected, held-out subset of the prompts of the training stimulus. Error bars show standard error of the mean (SEM) across training stimuli and initializations. (e) Change in mass of each attention category. (f) Change in correlation with human predictions and LM perplexity on unseen text. After optimization, human-model correlation increases after the first presentation of the stimulus (brown), but slightly decreases in the initial presentation (orange). Perplexity (blue), plotted here as the ratio of post- and pre-optimization performance, is hurt most in the middle layers.}
    \label{fig:attn_optim}
\end{figure*}

\section{Attention optimization}

Our previous results showed that human and LM next-word prediction accuracy diverge when short-term memory comes into play, suggesting that human and model memory mechanisms behave very differently. We then showed this divergence might be caused by the model's induction heads, which we hypothesized enable it to identify and recall patterns with superhuman accuracy.
We next asked if it is possible to modify the model so that its memory behaves more like the human.
Because the LM is superhuman, such a modification will selectively \emph{hurt} the LM's performance.

Since memory in this model is implemented through attention, we approached this problem by modifying the attention matrices of the model. We learn an additive bias $B_h$ for the attention matrix of each head $h$ in one layer such that adding this bias to the pre-softmax attention weights will produce outputs that are more human-like. Namely, we modify the attention mechanism in the model to be
\begin{equation}
    \text{Attn}(Q,K,V) = \text{softmax}\left(\frac{QK^T}{\sqrt{d}}\textcolor{red}{+B_h}\right)V
\end{equation}

Each stimulus consists of an $S$-token span presented $R$ times, for a total stimulus length $T=SR$.
Human and model top-1 accuracy for prompted word $i$ is denoted $P_{\text{human}}(\text{correct}_i)$ and $P_{\text{model}}(\text{correct}_i)$, respectively, and $N_i$ is the number of participants that responded to that prompt.
Let $B_h \in \mathbb{R}^{T \times T}$ be the additive bias for head $h$, and $H=12$ be the number of attention heads in each layer of GPT-2.
We optimize over $\{B_1, \dots, B_H\}$ to minimize the mean squared error (MSE) between $P_{\text{human}}(\text{correct})$ and $P_{\text{model}}(\text{correct})$, weighted by the number of subjects who responded to each prompt ($N_i$).
$W$ is the number of words that were prompted for at least one subject.
\begin{align}
    \min_{\{B_1, \dots, B_H\}} \frac{1}{W} \sum_{i=1}^{W} N_i \big( & P_{\text{human}}(\text{correct}_i) - \nonumber \\
    &P_{\text{model}}(\text{correct}_i) \big)^2
\end{align}

What form should $B_h$ take? 
The model is superhuman in its long-distance memory, so we sought to reduce the impact of long-distance attention by giving the model a recency bias.
Much earlier work has shown that human memory tends to decay as a power law with time~\cite{donkinPowerLawModelPsychological2012}.
A similar form of decay is also seen in mutual information between words as a function of their separation~\cite{linCriticalBehaviorPhysics2017}, and this has been previously exploited in designing efficient language models~\cite{mahtoMultitimescaleRepresentationLearning2020}.
To capture this type of behavior, we parameterized $B_h$ with $\alpha_h, \beta_h \in \mathbb{R}$:
\begin{equation}
    B_h = \sum_{k=0}^{T-1} \text{diag}_k(\alpha_{h} \cdot k ^ {-\exp( \beta_{h})})
\end{equation}
where $\text{diag}_{k}(d)$ constructs a $T \times T$ matrix that places the scalar $d$ along the $k$-th diagonal below the main diagonal.
Figure~\ref{fig:attn_optim}a shows an example matrix with this form.
This form of $B_h$ is advantageous because the effect of $\alpha_h, \beta_h$ can be evaluated on stimuli of any form or length, including those that are non-repeating.
We initialize $\alpha_h, \beta_h$ by sampling from a standard normal distribution.

We optimize the attention matrix biases $B_h$ to match human data from one stimulus over 2000 epochs via gradient descent with the Adam optimizer~\cite{kingmaAdamMethodStochastic2017}, and then evaluated human-model similarity with the other four stimuli.
For each training stimulus, we repeated this procedure with five initializations using different random seeds.
We set the learning rate to \num{5e-3}.

\subsection{Optimization results}
Because the long-range copying behavior seems to initiate in layer 6 (Figure~\ref{fig:attention}b), we began by only optimizing the attention bias for that layer.

We first examine the post-optimization timecourse of $P_{\text{model}}(\text{correct})$ by averaging the held-out accuracies for a single stimulus (Figure~\ref{fig:attn_optim}b).
While the model's predictions are largely unchanged in the initial presentation, performance significantly deviates toward human values in later presentations.
This is summarized in Figure~\ref{fig:attn_optim}c, where the model's average performance within the later presentations is closer to humans after optimization.
Importantly, this optimization procedure produces $B_h$ that generalize across stimuli because we do not fit on the human data for the held-out stimulus.

Additionally, these $B_h$ generalize \emph{within} the stimulus.
To measure within-stimulus generalization, we randomly selected 30\% of the prompts from each presentation of the span and calculated the MSE on this subset separately from the rest of the stimulus.
Figure~\ref{fig:attn_optim}d shows the training and held-out (validation) loss curves for the train stimulus, averaged across all five stimuli and five random initializations.
Training loss decreases on average $52.9\%$, while validation loss decreases $40.4\%$; most of the improvement for held-out prompts occurs in the first 1000 epochs.

We next examined the effects of the layer 6 intervention on the summarized attention patterns of each layer, similar to Figure~\ref{fig:attention}b.
Figure~\ref{fig:attn_optim}e shows the log-ratio of post- and pre-optimization probability mass for each attention pattern, averaged across all held-out stimuli.
The learned bias increases attention on the current token at the expense of all other measured patterns in layer 6, including (importantly) the induction pattern that would directly copy the correct token from a previous presentation.
Even though we only intervened in layer 6, the induction pattern is weaker in all following layers, and the model is attending more to the current and recent tokens.

Finally, we repeated the entire optimization procedure independently on each layer and evaluated the change in human-LM correlation.
We had hypothesized that our intervention should only work to create human-like behavior when applied to layers 6-12, which contained induction heads.
However, the intervention improved model-human correlation on repeated spans regardless of the layer on which optimization was performed (Figure~\ref{fig:attn_optim}f, brown line). Effects were strongest for layers 4-9, but small improvements were seen in every layer. This might suggest that induction heads are not the only important memory mechanism for this problem, or that the same effects can be achieved by modifying the inputs to induction heads.

Our results show that the recency bias intervention was effective at rescuing the divergence between human and model performance, but it is possible that this improvement comes at the cost of much worse model performance in other ways. For example, it could reduce the high correlation between human and model in scenarios lacking short-term memory, or make the model worse overall at next-word prediction. To test for the first effect, we computed the human-model correlation for the first presentation of each held-out stimulus (Figure~\ref{fig:attn_optim}f, orange line). We found that the correlation did fall, but by a much smaller amount than the correlation on subsequent presentations improved. For example, in layer 6 human-model correlation on the first presentation decreased by about 0.03, but the correlation on later presentations increased by 0.2.

We also tested whether our intervention increased LM perplexity on an unseen set of non-repeating text from the story corpus in order to measure how general LM abilities change due to the intervention.
No stories that were used for fine-tuning or constructing the repeating stimuli were used to measure perplexity. We computed the average perplexity for the modified and un-modified model, and reported their ratio (Figure~\ref{fig:attn_optim}f, blue line). We found that perplexity did increase due to the intervention, meaning that it generally harmed next-word prediction performance. However, the degree of increase varied substantially depending on which layer was modified, with the largest effect found in layer 6 (a more than $40\%$ increase) and smaller effects in the earliest and latest layers (roughly $10\%$ increase). This suggests that at least part of the model's general next-word prediction performance stems from its superhuman recall, and not its ability to mimic human cognition. Taking these three results together, we would suggest that the best layer to modify actually appears to be layer 9, which yields the largest improvement in human-model correlation with memory, a modest decline in human-model correlation without memory, and only a roughly $15\%$ increase in overall model perplexity.

\section{Conclusions}

Despite widely published results showing that human and LM prediction performance is comparable, we have found a scenario wherein humans and GPT-2 show a substantial divergence.
By examining the model's attention maps for non-initial presentations, we identify specific attention heads and layers that attend across presentation boundaries to copy the next token.
We finally demonstrate a procedure that augments these heads' attention maps with a recency bias, disrupting their copying behavior.
The intervention reliably improves human-LM similarity across held-out stimuli in later presentations, at the cost of increased perplexity.

With the behavioral data we collected, we have used an LM to build an explicit model of human memory.
Our findings here show that human memory has a stronger recency bias than GPT-2, and in the future we hope to use this model to learn more about human memory.
Additionally, it suggests that attending over long distances may result in diminishing returns---an alternate form of attention may be able to exploit this phenomenon for increased efficiency.

Further work must be done to describe the change in model states during repeated presentations of a stimulus.
Characterizing this experiment as a test of in-context learning (ICL), we may be able to exploit recent work \cite{daiWhyCanGPT2022} that suggests ICL is analogous to finetuning model weights.

\bibliography{states-vs-weights}
\bibliographystyle{acl_natbib}

\clearpage

\appendix

\section{Stimuli}
\label{sec:appendix_stimuli}

Below are the stimuli in their entirety.
Bolded words are those which at least one subject is asked to predict, given the previous ten words.
Presentation boundaries are marked with \textbf{//}, but this token is never presented to the subject or LM.

\noindent Stimulus 1 (3 presentations of a 65-word span):

\noindent\fbox{%
    \parbox{\linewidth}{%
        we start \textbf{to} \textbf{trade} stories about our lives we're both \textbf{from} up north we're \textbf{both} kind of \textbf{newish} to the neighborhood \textbf{this} is in \textbf{florida} we both went \textbf{to} college not great \textbf{colleges} but man \textbf{we} graduated and i'm \textbf{actually} finding myself a \textbf{little} \textbf{jealous} of her because she has this \textbf{really} cool job \textbf{washing} dogs she had horses back \textbf{home} and \textbf{she} really loves \textbf{//} we start \textbf{to} \textbf{trade} stories about our lives we're both \textbf{from} up north we're \textbf{both} kind of \textbf{newish} to the neighborhood \textbf{this} is in \textbf{florida} we both went \textbf{to} college not great \textbf{colleges} but man \textbf{we} graduated and i'm \textbf{actually} finding myself a \textbf{little} \textbf{jealous} of her because she has this \textbf{really} cool job \textbf{washing} dogs she had horses back \textbf{home} and \textbf{she} really loves \textbf{//} we start \textbf{to} \textbf{trade} stories about our lives we're both \textbf{from} up north we're \textbf{both} kind of \textbf{newish} to the neighborhood \textbf{this} is in \textbf{florida} we both went \textbf{to} college not great \textbf{colleges} but man \textbf{we} graduated and i'm \textbf{actually} finding myself a \textbf{little} \textbf{jealous} of her because she has this \textbf{really} cool job \textbf{washing} dogs she had horses back \textbf{home} and \textbf{she} really loves
    }
}

\noindent Stimulus 2 (3 presentations of a 61-word span):

\noindent\fbox{%
    \parbox{\linewidth}{%
        get out \textbf{to} the \textbf{hamptons} and we're at this \textbf{farmhouse} and it \textbf{was} like a \textbf{scene} out of christopher \textbf{isherwood} \textbf{the} berlin stories all these blonde \textbf{boys} about ten of us \textbf{running} \textbf{around} doing \textbf{push} ups so that our \textbf{muscles} would swell and in and \textbf{out} of the pool \textbf{and} a big buffet and everything \textbf{waiting} for \textbf{the} light to change \textbf{//} get out \textbf{to} the \textbf{hamptons} and we're at this \textbf{farmhouse} and it \textbf{was} like a \textbf{scene} out of christopher \textbf{isherwood} \textbf{the} berlin stories all these blonde \textbf{boys} about ten of us \textbf{running} \textbf{around} doing \textbf{push} ups so that our \textbf{muscles} would swell and in and \textbf{out} of the pool \textbf{and} a big buffet and everything \textbf{waiting} for \textbf{the} light to change \textbf{//} get out \textbf{to} the \textbf{hamptons} and we're at this \textbf{farmhouse} and it \textbf{was} like a \textbf{scene} out of christopher \textbf{isherwood} \textbf{the} berlin stories all these blonde \textbf{boys} about ten of us \textbf{running} \textbf{around} doing \textbf{push} ups so that our \textbf{muscles} would swell and in and \textbf{out} of the pool \textbf{and} a big buffet and everything \textbf{waiting} for \textbf{the} light to change

    }
}

\newpage %

\noindent Stimulus 3 (3 presentations of a 52-word span):

\noindent\fbox{%
    \parbox{\linewidth}{%
        nine \textbf{hours} i find myself \textbf{nine} hours later back in the \textbf{situation} \textbf{room} looking through \textbf{the} glass window \textbf{at} the operations people hoping this works when i see people start \textbf{cheering} and \textbf{erupting} in cheers and excited and i \textbf{hear} alice \textbf{bowman's} voice over the \textbf{intercom} we \textbf{are} back \textbf{on} the prime \textbf{//} nine \textbf{hours} i find myself \textbf{nine} hours later back in the \textbf{situation} \textbf{room} looking through \textbf{the} glass window \textbf{at} the operations people hoping this works when i see people start \textbf{cheering} and \textbf{erupting} in cheers and excited and i \textbf{hear} alice \textbf{bowman's} voice over the \textbf{intercom} we \textbf{are} back \textbf{on} the prime \textbf{//} nine \textbf{hours} i find myself \textbf{nine} hours later back in the \textbf{situation} \textbf{room} looking through \textbf{the} glass window \textbf{at} the operations people hoping this works when i see people start \textbf{cheering} and \textbf{erupting} in cheers and excited and i \textbf{hear} alice \textbf{bowman's} voice over the \textbf{intercom} we \textbf{are} back \textbf{on} the prime
    }
}

\noindent Stimulus 4 (2 presentations of a 107-word span):

\noindent\fbox{%
    \parbox{\linewidth}{%
        year during the \textbf{seventies} my four \textbf{aunts} \textbf{would} take me \textbf{and} my two cousins on their dream vacation a \textbf{rented} beach house in \textbf{hyannis} \textbf{on} the very \textbf{cove} sharing \textbf{beachfront} with the kennedy \textbf{compound} every day for an entire week my aunt pat would \textbf{roll} up her \textbf{sisters'} hair my aunts \textbf{would} apply \textbf{sunscreen} \textbf{to} the \textbf{back} of their \textbf{necks} the backs of \textbf{the} hands \textbf{and} the \textbf{tops} of their feet and \textbf{then} they would drag their beach chairs down to \textbf{the} beach and they would set them up perfectly \textbf{not} facing the water not into the \textbf{sun} for tanning but perfectly \textbf{for} \textbf{spying} on \textbf{the} kennedys \textbf{//} year during the \textbf{seventies} my four \textbf{aunts} \textbf{would} take me \textbf{and} my two cousins on their dream vacation a \textbf{rented} beach house in \textbf{hyannis} \textbf{on} the very \textbf{cove} sharing \textbf{beachfront} with the kennedy \textbf{compound} every day for an entire week my aunt pat would \textbf{roll} up her \textbf{sisters'} hair my aunts \textbf{would} apply \textbf{sunscreen} \textbf{to} the \textbf{back} of their \textbf{necks} the backs of \textbf{the} hands \textbf{and} the \textbf{tops} of their feet and \textbf{then} they would drag their beach chairs down to \textbf{the} beach and they would set them up perfectly \textbf{not} facing the water not into the \textbf{sun} for tanning but perfectly \textbf{for} \textbf{spying} on \textbf{the} kennedys
    }
}

\newpage %

\noindent Stimulus 5 (4 presentations of a 57-word span):

\noindent\fbox{%
    \parbox{\linewidth}{%
        pastor \textbf{was} this forty something \textbf{british} guy and he really wanted \textbf{to} \textbf{attract} \textbf{twenty} \textbf{somethings} so we were a hot \textbf{commodity} we \textbf{were} right in the demographic and we \textbf{started} to get promoted up into \textbf{higher} and higher \textbf{echelons} of leadership so we were invited \textbf{to} the \textbf{leadership} team meeting and then the \textbf{core} \textbf{leadership} team meeting \textbf{//} pastor \textbf{was} this forty something \textbf{british} guy and he really wanted \textbf{to} \textbf{attract} \textbf{twenty} \textbf{somethings} so we were a hot \textbf{commodity} we \textbf{were} right in the demographic and we \textbf{started} to get promoted up into \textbf{higher} and higher \textbf{echelons} of leadership so we were invited \textbf{to} the \textbf{leadership} team meeting and then the \textbf{core} \textbf{leadership} team meeting \textbf{//} pastor \textbf{was} this forty something \textbf{british} guy and he really wanted \textbf{to} \textbf{attract} \textbf{twenty} \textbf{somethings} so we were a hot \textbf{commodity} we \textbf{were} right in the demographic and we \textbf{started} to get promoted up into \textbf{higher} and higher \textbf{echelons} of leadership so we were invited \textbf{to} the \textbf{leadership} team meeting and then the \textbf{core} \textbf{leadership} team meeting \textbf{//} pastor \textbf{was} this forty something \textbf{british} guy and he really wanted \textbf{to} \textbf{attract} \textbf{twenty} \textbf{somethings} so we were a hot \textbf{commodity} we \textbf{were} right in the demographic and we \textbf{started} to get promoted up into \textbf{higher} and higher \textbf{echelons} of leadership so we were invited \textbf{to} the \textbf{leadership} team meeting and then the \textbf{core} \textbf{leadership} team meeting
    }
}

\section{Additional GPT-2 experiments}
\label{sec:appendix_spans}
Our human-LM comparisons were limited by the amount of data we could collect from our behavioral experiment, but GPT-2 has no such limitation.
We further tested the LM on 100 random, non-phrase-aligned spans of text of different lengths (10 to 570 words, in increments of 40) from the corpus of annotated spoken narratives~\cite{lebelNaturalLanguageFMRI2023}.
For each text span, we form a stimulus by repeating the span 15 times, or until the resulting text exceeds the maximum input length of the model -- in this case, 1024 tokens for GPT-2.

We feed each stimulus into the model and calculate the perplexity for every token in the input.
For each span length, we average the perplexity across the 100 random spans, yielding a single perplexity measure per token position.
We finally average the perplexity within the tokens of each presentation.

\subsection{Results}

\begin{figure*}
	\centering
	\includegraphics[width=\textwidth]{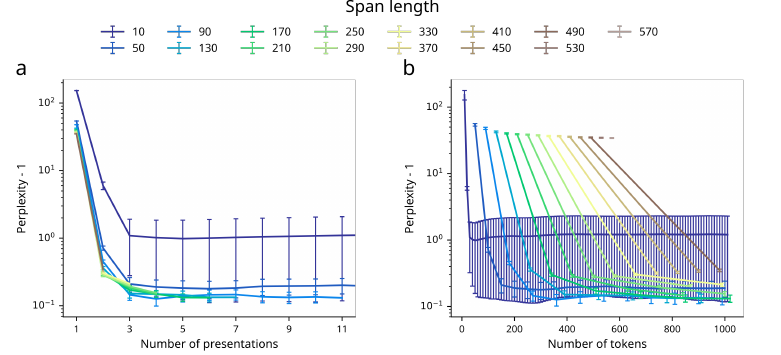} 
	\caption{Model results for GPT-2. (a) shows the average perplexity for each presentation. (b) changes the x-axis to show the total number of tokens.}
	\label{fig:model_gpt2}
\end{figure*}

Figure \ref{fig:model_gpt2} shows results for the repeated span experiment for GPT-2.
GPT-2's perplexity on the initial presentation improves with longer spans. After only a few presentations, however, the perplexity for GPT-2 quickly plateaus to near-perfect performance. The model effectively memorizes the span, and has learned when to regurgitate the previously seen tokens.
These results confirm the observations in Figure~\ref{fig:basic_results} on a significantly larger set of stimuli.
For smaller spans at higher repeats, though the mean perplexity across spans remains stable with more presentations, the standard deviation increases substantially. %

These results extend the findings for LMs in Figure~\ref{fig:basic_results} to more presentations.

\end{document}